# Synthesizing time-series wound prognosis factors from electronic medical records using generative adversarial networks

Farnaz H. Foomani[1], D. M. Anisuzzaman[2], Jeffrey Niezgoda[3], Jonathan Niezgoda[3], William Guns[3], Sandeep Gopalakrishnan[4,*], Zeyun Yu[1,2,*]

[1]Department of Electrical Engineering, University of Wisconsin-Milwaukee, Milwaukee, WI, United States
[2]Department of Computer Science, University of Wisconsin-Milwaukee, Milwaukee, WI, United States
[3]AZH Wound and Vascular Center, Milwaukee, WI, United States
[4]College of Nursing, University of Wisconsin-Milwaukee, Milwaukee, WI, United States

[*]Corresponding authors:
Zeyun Yu, Email: yuz@uwm.edu
Sandeep Gopalakrishnan, Email: sandeep@uwm.edu

## Abstract

Wound prognostic models not only provide an estimate of wound healing time to motivate patients to follow up their treatments but also can help clinicians to decide whether to use a standard care or adjuvant therapies and to assist them with designing clinical trials. However, collecting prognosis factors from Electronic Medical Records (EMR) of patients is challenging due to privacy, sensitivity, and confidentiality. In this study, we developed time series medical generative adversarial networks (GANs) to generate synthetic wound prognosis factors using very limited information collected during routine care in a specialized wound care facility. The generated prognosis variables are used in developing a predictive model for chronic wound healing trajectory. Our novel medical GAN can produce both continuous and categorical features from EMR. Moreover, we applied temporal information to our model by considering data collected from the weekly follow-ups of patients. Conditional training strategies were utilized to enhance training and generate classified data in terms of healing or non-healing. The ability of the proposed model to generate realistic EMR data was evaluated by TSTR (test on the synthetic, train on the real), discriminative accuracy, and visualization. We utilized samples generated by our proposed GAN in training a prognosis model to demonstrate its real-life application. Using the generated samples in training predictive models improved the classification accuracy by 6.66-10.01% compared to the previous EMR-GAN. Additionally, the suggested prognosis classifier has achieved the area under the curve (AUC) of 0.975, 0.968, and 0.849 when training the network using data from the first three visits, first two visits, and first visit, respectively. These results indicate a significant improvement in wound healing prediction compared to the previous prognosis models.

***Index Terms:*** *Generative adversarial network, Wound prognosis model, Electronic medical record*

## Introduction

It is reported that more than 6 million people in the United States are suffering from various types of chronic wounds; Venous Leg Ulcer (VLU), Arterial Ulcer (AU), Diabetic Foot Ulcer (DFU), and Pressure Ulcer (PU), and annually more than $25 billion is spent on wound management and Medicare cost (1). About 0.15% to 0.3% of people are suffering from active VLU worldwide. Although there is no consensus about wound healing time, a wound is considered chronic if it has not healed in 4-12 weeks or

has shown less than 20 % reduction in its area after a maximum of four weeks of treatment (2). An accurate estimate of healing time could assist clinicians in making better decisions about treatments and interventions. This study aims to predict the wound healing status of VLU patients based on data collected from the first three visits. We developed a deep learning model to predict the VLU healing status within 12 weeks after receiving the initial treatment. It has been reported that the best prognostic factors come from weekly follow-ups and evaluation of the wound healing process for a minimum of three weeks (3). Therefore, we used data from the electronic medical record (EMR) of patients diagnosed with VLU on their first three weekly visits.

In the past decade, machine learning (ML) techniques have been extensively used for disease diagnosis, such as kidney disease (4), skin cancer (5), breast cancer (6), heart disease (7), and retinal layer segmentation to diagnose Alzheimer's disease (8). However, little work has been accomplished in chronic wound healing prediction. Conventional ML algorithms such as logistic regression, support vector machine, decision trees, and random forests are highly dependent on feature representations. Moreover, the shallow models are incapable of simulating the complexity of decision-making of a human neuronal system (9). On the other hand, deep models are inspired by the multi-level cognition of the human brain and have proved to be able to model the nonlinearity and complexity of human thinking. Deep neural network models have the potential of automatic feature extraction and can abstract high-level representations from low-level information (10). Therefore, in this study, a prognosis convolutional neural network (Prog-CNN) was trained to predict the healing rate of patients with VLU within 12 weeks of receiving their first treatment. To the best of our knowledge, it is the first time that deep learning models have been used for wound healing prediction.

EMR is an important resource to help clinicians in diagnosing the onset or predicting the future condition of a specific disease (11). EMR of a patient includes coded diagnoses, interventions, and treatments as well as some text information that usually contains precious clinical data about the specific visit (12). Recently, studies are focused on extracting features from EMR and developing ML models to assist clinicians in diagnosis and prediction. However, due to the sensitivity and confidentiality of EMR, obtaining access to data is limited.

To overcome EMR data access challenges, an alternative method has been created to generate synthetic instances that look realistic using generative adversarial networks (GANs). GANs received much attention recently because of their ability to produce high-quality synthetic images. In GANs, two neural networks are deployed; the first network is a *generator* that is trained to produce realistic instances from the latent space which can mislead the second network (*discriminator*) into identifying them as the original data. Recently, a few studies have been performed on medical records to produce synthetic structured and categorical data (13-16).

The present study aims to create more realistic synthetic EMR data by incorporating deep learning into EMR-Conditional Wasserstein Generative Adversarial Network (EMR-CWGAN). In our proposed GAN, EMR-Time-series Conditional Wasserstein Generative Adversarial Network (EMR-TCWGAN), the generator and discriminator (called the critic in this work) are built from two convolutional neural networks. Wasserstein GAN with gradient penalty (WGAN-gp), and the conditional GAN strategy were applied to avoid the vanishing gradients and mode collapsing (17-19). Finally, the ability of the suggested model in producing realistic data was tested by training Prog-CNN on the synthetic data and testing on the real data.

## Related Work

A review of the literature related to wound progress was conducted and the highlighted related works are presented. We will discuss the factors upon which the prognostic models have been developed and review the recent advances in chronic wound prognosis models based on machine learning techniques. Finally, we summarize specific investigations in medical generative adversarial networks (GANs) to generate synthetic EMR.

*Wound prognosis factors:* Patients who have four weeks of prognostic information provided to a clinic are more likely to heal than patients without any prognostic records. Shanu K. Kurd et al. (20) suggested that the most important factor associated with a healed wound is the change in wound size after four weeks of care. Skene et al. (21) stated smaller initial ulcer area, shorter duration of ulceration, younger age, and no involvement of deep vein as the most significant healing predictors of VLU. Franks et al. (22) reported ulcer size, ulcer duration, general mobility, and limb joint mobility as the important predictors of leg ulceration healing. Vesna Karanikolic et al. (23) declared that factors associated with the delay of VLU healing are infection, number, and larger surface area of ulcer. Ankle-brachial pressure index and the presence of lipodermatosclerosis are important positive factors for the healing of VLU. Factors that do not appear to have significant roles in healing are age, sex, obesity, condition of the surface, the deep venous system, and chronic diseases. Wound size and duration are the most important factors in developing a prognostic model for VLU. In addition to these two factors, ulcer grade and wound number are some important characteristics for developing the model (24). The most significant predictors for delayed healing of neuropathic diabetic foot ulcer (DFU) as stated by Margolis et al. (25) are wound size, duration, and ethnicity of patients. Significant predictors mentioned by the same authors for VLU are wound area and duration, ankle-brachial index, ethnicity, limb ulcer, history of stripping or venous ligation, inability to walk (1 block), wound margin, lipodermatosclerosis, fibrin-covered wound, and history of surgical wound debridement (26). Khachemoune et al. (27) have asserted that wound size and duration, lipodermatosclerosis, and history of failed prior split-thickness skin grafts are the main factors impacting healing chronic VLUs, treated with Cryopreserved Epidermal Cultures (CEC). Ulcer duration and area are the most significant factors influencing the healing of VLUs; where patient sex, age, race, skin condition, and infection have no prognostic significance (28). In summary, the most common significant wound healing predictors are wound size (length, surface area), depth, grade, duration, distance, color, and numbers; ankle brachial index; ethnicity; lipodermatosclerosis; and previous wound treatment history. The most common non-significant wound healing predictors are the patient's age and sex, body mass index (obesity), the deep venous system, and infection.

*Wound prognosis models:* In 2020, Cho et al. (24) reported a wound healing predictive model based on logistic regression and classification tree models and achieved an area under the curve (AUC) of 0.712 and 0.717 respectively by training the models using the dataset collected from the first intake visit. Their dataset included AU, DFU, PU, and VLU and their relative wound measurements, as well as patients' clinical and demographic characteristics. To avoid model complexity and overfitting, they added variables in a stepwise fashion to evaluate their contribution to the model performance considering the AUC as their metrics. Jung et al. (25) developed their model using logistic regression, random forest and gradient boosted tree models and reported the AUC between 0.834 and 0.847. The training dataset consisted of information including patient age, sex, insurance, and zip codes, wound information including 40 different wound types and 37 wound locations, as well as wound assessments such as dimension, edema, erythema, and rubor. Data from the first two weeks of the wound assessment was used in training the network. Cukjati et al. (3) performed a classification decision tree to predict the wound healing rate after one to six weeks of follow-up. The data included three categories: 1) Wound characteristics (length, width, depth, grade, date of appearance, date of treatment beginning, etiology, and location), 2) Patient characteristics (sex, age, number of wounds, diagnosis, date of spinal cord injury, and degree of spasticity), and 3) Treatment/management (type of treatment, daily duration of treatment, duration of treatment). The classification accuracy reported with data available for two weeks was 62% and, for three weeks of follow-up, it increased to 80%. Margolis et.al (29) built their prognostic model using logistic regressions based on different numbers of DFU prognosis variables to predict the wound status by the $20^{th}$ week of care. Using variables such as age, sex, and the number, duration, size and grade of wounds, they achieved a maximum AUC of 0.70.

*Medical generative adversarial networks:* MedGAN was introduced by Choi, et al. (15) in 2017 to generate high-dimensional discrete variables by incorporating an autoencoder in generative adversarial networks. Diagnosis, medication, and procedure codes in EHR data were expressed as a vector, where the $i^{th}$ array indicates the number of occurrences of the $i^{th}$ variable in a patient (counts). Moreover, a binary vector representation of the EHR data was performed, in which the $i^{th}$ dimension can be represented by 0 or 1, indicating the absence or presence of the $i^{th}$ variable in a patient's record. In MedGAN, autoencoders learn from the count and binary discrete input vectors to map them to a lower dimensional space and reconstruct the original input in the output by mapping them back to the original dimension. The generator needs to learn this low dimensional representation of the original data; therefore, the same decoder was used after the generator to reconstruct the original dimension. Later, in 2018, Baowaly, et al.(14) integrated the idea of the Wasserstein GAN with gradient penalty and boundary seeking GAN to generate more realistic synthetic patient records by introducing medWGAN and medBGAN. They performed K–S similarity test and reported a 3% of improvement in the quality of the generated data using medBGAN compared to the medGAN. , Zhang et al. (16) reported EMR-WGAN and EMR-CWGAN in which the autoencoder was removed due to the model bias. Moreover, they introduced a conditional training strategy and incorporated the concept of labels as part of the generator and discriminator.

Although medical generative adversarial networks have recently shown acceptable performance in generating synthetic patients' records, there have been some limitations which have been taken into account in this study:

1. The previous studies have only focused on binary features and the input dataset only included count and binary representation of the medical record. Moreover, the size of the input vector is equal to the number of the diagnosis, medication, and procedure codes in the available EHR data. This representation not only creates a very high dimensional input vector, but more importantly it also requires having access to a large EMR dataset. With a small dataset the GAN will generate a very limited distribution of patient's records as it is not possible to include all the existing diagnoses, medications, and procedure codes in the dataset. To handle these limitations, our dataset consists of both categorical and continuous features. The size of the input data equals the number of the wound prognosis factors which are listed in Table 1. To reduce the size of the input data, instead of training an autoencoder, a random forest regression model was trained to select the variables which have greater influence on the healing process. With this method the generator needs to learn to produce only the informative features which make the model less complex, and this representation of the medical records does not require a big dataset thus overcoming the challenges of access to huge EMR/EHR data.

2. Previous studies have used a statical EHR dataset and the temporal information, which represents a disease evolution that has not been considered. For example, in wound assessments, the variations in the wound dimensions are considered as an important factor to predict the healing status. In order to generate a realistic synthetic medical record, time will be included and modeled accordingly in the current study.
3. The data generated by previous medGANs has never been applied in a real-life application. This evaluation is important to assess the quality of the generated data as well as their practicality. Hence, a wound prognosis model was trained in the present work to predict a chronic wound healing status within 12 weeks of the initial intake exam. The model was trained using the generated instances and tested on the original EMR to investigate the functionality of our synthetic medical records.

## Material and Method

This section provides a description of the structure of the EMR data we adopted in this work followed by a detailed description of our proposed EMR-TCWGAN and Prog-CNN.

### EMR Dataset

The data in this study is derived from the EMR of patients diagnosed with VLU in AZH Wound and Vascular Centers, Milwaukee, WI. The data has been carefully de-identified and includes patients' general information such as age, sex, ethnicity, their wound-related information such as wound length, width, area, location, and duration as well as their clinical information including the history of any vascular diseases, systolic and diastolic blood pressure, BMI, etc. The data contains both categorical and continuous values. Some of the features such as age, BMI, systolic and diastolic blood pressures were discretized and converted to the categorical form. However, wound measurements, including wound length, width, and area remained continuous. Table 1 represents the prognosis factors that were included in this study along with their categories and statistics in detail.

Each patient went through a weekly follow-up up to 12 weeks and their wounds were evaluated at the end of each visit and labeled as healed or not healed by the expert physician. Generally, a wound is considered healed if the wound has zero measurements (30). However, the ground truth labels are the status of a wound by week 12 of the first visit. Our initial data includes the medical records of 70 patients. Patients with less than 3 weeks of wound assessments were excluded from the dataset. Those who stopped follow-ups before 12 weeks were considered not healed unless the termination was due to their healing in less than 12 weeks. Since not all the patients followed a weekly visit regularly, we applied this irregularity by defining a new parameter, *separator,* to indicate the time gaps between each visit.

Table 1-Summary statistics of EMR dataset

| Prognosis factor | Percentage of prognosis factors | | | Prognosis factor | Percentage of prognosis factors | | |
|---|---|---|---|---|---|---|---|
| | 1st visit | 2nd visit | 3rd visit | | 1st visit | 2nd visit | 3rd visit |
| **Age** | | | | **Systolic blood pressure** | | | |
| <55 | 8.69 | 8.69 | 8.69 | <120 | 6.52 | 10.87 | 8.69 |
| 56-64 | 15.22 | 15.22 | 15.22 | 121-129 | 13.04 | 2.17 | 8.69 |
| 65-74 | 32.60 | 32.60 | 30.43 | 130-139 | 26.08 | 17.39 | 21.73 |
| >75 | 43.48 | 43.48 | 45.65 | 140-179 | 41.30 | 58.69 | 54.35 |
| | | | | >180 | 13.04 | 10.87 | 6.52 |
| **Sex** | | | | **Diastolic blood pressure** | | | |
| Female | | 54.34 | | <80 | 76.09 | 86.95 | 84.78 |
| Male | | 45.65 | | 81-89 | 15.22 | 8.69 | 10.86 |
| | | | | 90-119 | 8.69 | 4.35 | 4.34 |
| | | | | >120 | 0.0 | 0.0 | 0.0 |
| **Ethnicity** | | | | **Wound location** | | | |
| Black | | 15.22 | | Mid leg | | 67.39 | |
| White | | 84.78 | | Distal | | 19.56 | |
| others | | 0.0 | | Anterior leg | | 4.34 | |
| | | | | Lateral leg | | 6.52 | |
| | | | | others | | 2.17 | |
| **Smoking status** | | | | **Wound duration** | | | |
| Yes | | 8.69 | | Less than 4 weeks | 47.82 | 39.13 | 23.91 |
| No | | 43.47 | | 1-3 months | 23.91 | 32.61 | 47.82 |
| Reformed | | 47.82 | | Greater than 3 months | 28.26 | 28.26 | 28.26 |
| **BMI** | | | | **Prior wound infection** | | | |
| <18.5 | 0.0 | 0.0 | 0.0 | Yes | | 10.87 | |
| 18.5-24 | 4.35 | 4.35 | 6.52 | No | | 82.61 | |
| 25-29 | 34.78 | 34.78 | 32.60 | Unknown | | 6.52 | |
| >30 | 60.87 | 60.87 | 60.87 | | | | |
| **History of DVT** | | | | **Prior ulcer grafting** | | | |
| Yes | | 13.04 | | Yes | | 0.0 | |
| No | | 84.78 | | No | | 97.83 | |
| Unknown | | 2.17 | | Unknown | | 2.17 | |
| **History of VD** | | | | **Percentage of wound covered with fibrin** | | | |
| Yes | | 8.69 | | <25% | 50.0 | 58.69 | 58.69 |
| No | | 91.30 | | 25-50% | 8.69 | 6.52 | 4.35 |
| | | | | 50-75% | 13.04 | 13.04 | 13.04 |
| | | | | 75-100% | 6.52 | 15.21 | 17.39 |
| | | | | 100% | 21.74 | 6.52 | 6.52 |
| **Diabetes Type** | | | | **Percentage of wound covered with eschar** | | | |
| Insulin | | 0.0 | | <25% | 86.95 | 82.61 | 86.95 |
| Oral medications or diet | | 54.35 | | 25-50% | 4.35 | 6.52 | 2.17 |
| No diabetes | | 45.65 | | 50-75% | 2.17 | 4.35 | 4.34 |
| | | | | 75-100% | 0.0 | 0.0 | 0.0 |
| | | | | 100% | 6.52 | 6.52 | 6.52 |

continue

| Prognosis factor | Percentage of prognosis factors | | | Prognosis factor | Percentage of prognosis factors | | |
|---|---|---|---|---|---|---|---|
| | 1st visit | 2nd visit | 3rd visit | | 1st visit | 2nd visit | 3rd visit |
| **Drainage** | | | | **Doppler pulses** | | | |
| None | 4.35 | 2.17 | 21.74 | None | 56.52 | 54.35 | 54.35 |
| Mild | 36.96 | 60.87 | 43.48 | Monophasic | 21.74 | 23.91 | 23.91 |
| Moderate | 56.52 | 34.78 | 32.61 | Biphasic | 17.40 | 17.39 | 17.39 |
| Heavy | 2.17 | 2.17 | 2.17 | Triphasic | 4.35 | 4.35 | 4.35 |
| **Edema** | | | | **Doppler evidence insufficiency** | | | |
| None | | 2.17 | | No | 69.56 | 43.48 | 30.43 |
| Mild | | 28.26 | | GSV/SSV | 26.09 | 52.17 | 65.22 |
| Moderate | | 60.86 | | Deep | 0.0 | 0.0 | 0.0 |
| Heavy | | 8.69 | | Both | 97.83 | 4.35 | 4.35 |
| **Treatment** | | | | **CEAP wound stage** | | | |
| 20-30/30-40 mmHg Compression Stockings | 82.61 | 84.78 | 82.61 | No visual or palpable signs of CVD | 0.0 | 0.0 | 0.0 |
| 3-4- Layer Compression | 80.43 | 80.43 | 73.91 | Telangiectasia or reticular veins | 0.0 | 0.0 | 0.0 |
| Edema ware/Farrow wrap/Spandagrips and Short Stretch Bandage | 13.04 | 19.56 | 21.74 | Varicose veins | 0.0 | 0.0 | 0.0 |
| | | | | Edema | 0.0 | 0.0 | 0.0 |
| | | | | Pigmentation: skin changes - hemosiderin staining | 2.17 | 0.0 | 0.0 |
| Pneumatic Pump | 0.0 | 0.0 | 0.0 | Healed ulcer | 0.0 | 0.0 | 19.56 |
| Sharp/ultrasonic Debridement | 0.02 | 0.06 | 0.0 | Active ulcer | 97.83 | 100 | 80.43 |
| Central Venous treatment/Peripheral Venous Ablation | 0.0 | 0.04 | 0.04 | | | | |
| **Sharp debridement** | | | | **Endogenous intervention** | | | |
| Yes | 8.69 | 13.04 | 8.69 | Yes | 10.87 | 15.22 | 17.39 |
| No | 91.31 | 86.95 | 91.31 | No | 89.13 | 84.78 | 82.61 |
| **DermaPace** | | | | **HBOT** | | | |
| Yes | | 0.0 | | Yes | 2.17 | 2.17 | 0 |
| No | | 100 | | No | 97.82 | 97.82 | 100 |
| **Separator** | | | | **Lipodermatosclerosis** | | | |
| 1week | | 91.30 | 86.95 | Yes | | 39.13 | |
| 2weeks | NAN | 2.17 | 10.86 | No | | 60.87 | |
| >3weeks | | 6.52 | 2.17 | | | | |

To handle the missing data, we applied polynomial regressions in continuous and averaging in the categorical variables. The final dataset included the medical records of 60 patients, in which 75% of data (45 patients) were randomly selected as the training dataset and the remaining data (15 patients) were used as the test set.

Previous studies have reported that wound level factors such as wound dimensions and locations would result in substantial enhancement in predictive accuracy (30, 31). Besides the wound measurements, clinical variables that show a changing rate on a weekly basis such as wound fibrin and eschar percentage have more prognostic values in time series analysis when compared to those that are fixed between each visit such as sex and ethnicity. Thus, in this study, the analysis of the relative importance of variables has been conducted based on the Random Forest (RF) regression to identify the factors with more prognostic information. Note that RF regression has been trained using the first assessment for each patient to avoid any looking into the future.

## Preliminary: Generative Adversarial Network

GANs were first introduced by Ian J. Goodfellow, et.al, in 2014 (32). The main idea is to train two networks, generator "G" and discriminator "D", simultaneously. The generator learns the distribution of the data and outputs a sample that looks like real data. On the other hand, the discriminator, which is a binary classifier, needs to classify the sample as real or fake. In this minimax two-player game, G maps an input noise, $p_z$, to data space, and D is trained to maximize the probability of accurate classification of real and fake samples: $\log D(x)$. To produce more realistic samples, G is trained to minimize the difference between the discriminator output and real labels: $\log(1 - D(G(z)))$. Formally, the game between G and D is represented by the following equation in which $p_{data}$ is data distribution, and $p_z(z)$ is a Gaussian noise distribution :

$$\min_{G}\max_{D} V(D,G) = E_x \sim p_{data}(x)[\log[D(x)] + E_z \sim p_z(z)[\log(1 - D(x)] \tag{1}$$

Later, Arjovsky et al. (17) introduced WGAN in which Jensen-Shannon (JS) divergence was replaced with Wasserstein divergence. It is reported that WGANs could solve the main training challenges of GANs such as the requirement of maintaining balance in training the generator and discriminator, dependency of the network's architecture, and overcoming mode drop (failure in generating all the underlying distribution of the original data) and mode collapse (generation same output from different inputs).

## EMR Timeseries Conditional Wasserstein GAN (EMR-TCWGAN)

In the proposed EMR-TCWGAN which is illustrated in Figure 1, we employed WGAN-GP introduced by Ishaan Gulrajani et. al to minimize the optimization difficulties that occur occasionally in weight clipping, by penalizing the norm of the gradient of the critic with respect to its input (18). We utilized the conditional training strategy, in which we used prognosis labels, healed vs. not healed, to generate labeled synthetic data useful in training our prognosis network. We incorporated the prognosis labels into the generator and critic.

The generator network, G, is a CNN, which takes the input noise and the desired label (healed vs. not healed) and outputs the value of the time series which is a $T_x$ by $n_x$ matrix in which $T_x$ is the number of successive visits and $n_x$ is the number of prognosis variables. This transformation is done through a dense layer with $(T_x/2) * (n_x/2) * 256$ neurons, and three deconvolution layers with LeakyReLU activation functions, batch normalization, and dropout at each layer except for the last one in which the activation function is Tanh. The number of 3 by 3 filters are 128, 64, and 1 for each layer. The time dimension is zero-padded for an odd number of $T_x$.

The critic, C, is a CNN, which takes real data or a generated data, and their associated true labels. The true labels have the same dimension as the real and generated data. C outputs a score, representing whether the data is real or generated. The critic network is composed of four layers of convolution with LeakyReLU activation functions followed by a dropout layer. The last layer is a dense layer with one neuron and a linear activation function. The number of 3 by 3 filters for each convolution layer is 64, 128, 256, and 512.

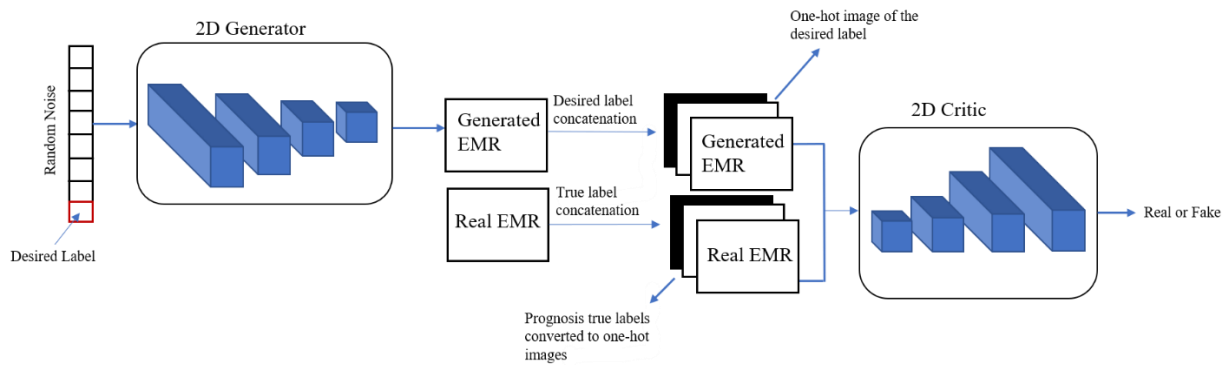

Figure 1- Architecture of EMR-TCWGAN

## Prognosis convolutional neural network (Prog-CNN)

We used a CNN model to predict the probability of a wound being healed by the end of week 12 as a function of our selected prognosis factors. The Prog-CNN was trained on both train and synthetic datasets and tested on the test dataset. This classifier is constructed of two convolution layers followed by a dropout layer. There are 16 filters of size 3 x 3 for each layer. In the end, there are two fully connected layers with 5 and 1 neurons having a sigmoid activation function. Figure 2 illustrates the developed Prog-CNN architecture.

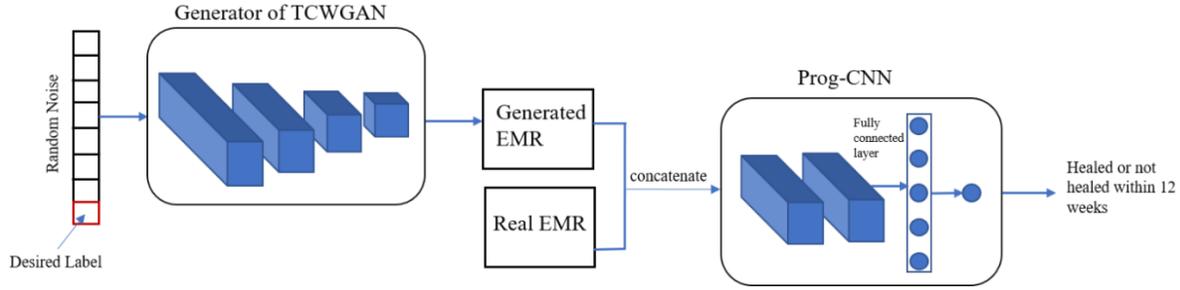

Figure 2-Architecture of Prog-CNN trained by synthetic and real data.

# Results

The analysis of the relative variable importance based on the random forest regression model is shown in Figure 3. Wound length had the highest variable importance in predicting wound healing, and the importance of the other variables is presented, respectively. All three wound measurements; wound length, width, and area, are considered the most important variables for healing prediction with a score of 1.00, 0.83, and 0.67, respectively. Doppler evidence and fibrin percentage are listed as the second and the third important variables in prediction. Surprisingly, the least important predictors were separator, which represents the irregular visits of a patient. After that, DermaPACE and CEAP wound are listed as the less important variables. Although having more prognosis factors could increase the predictive power of the model, it may cause complexity and overfitting (30). Therefore, to increase the predictive accuracy, we disregarded variables with relative importance less than 0.3. Hence, we ended up with a total of 14 predictive features including, wound length, width, area, Doppler evidence, percentage of fibrin, the number of Doppler pulses, systolic blood pressure, age, duration of the wound, history of DVT, wound location, drainage, diabetes type, and edema. All the 14 predictive features were used to train both EMR-TCWGAN, and the Prog-CNN.

## Evaluation of EMR-TCWGAN

We compare our proposed model (EMR-TCWGAN) with EMR-CWGAN, the most recent and related method to ours. To assess the quality of the generated data three criteria are considered; (1) the distribution of the generated data should be matched with the real data (33), (2) samples should be as much useful as the original data in real-life applications (Train on Synthetic, Test on Real; TSTR) (34), (3) the generated samples should not be distinguishable from the original data (33).

(1) *Visualization with t-SNE:* t-SNE was applied on both synthetic and original data (Test and Train) for visualizations in two dimensions. Before applying t-SNE, the temporal dimension was flattened. This evaluation can show the similarity of the distribution of the original and synthetic data. We applied t-SNE on the test dataset to assess the ability of EMR-TCWGAN to generate synthetic data that can cover the distribution of unseen data points as well. Note that the test dataset was not used in training EMR-TCWGAN.

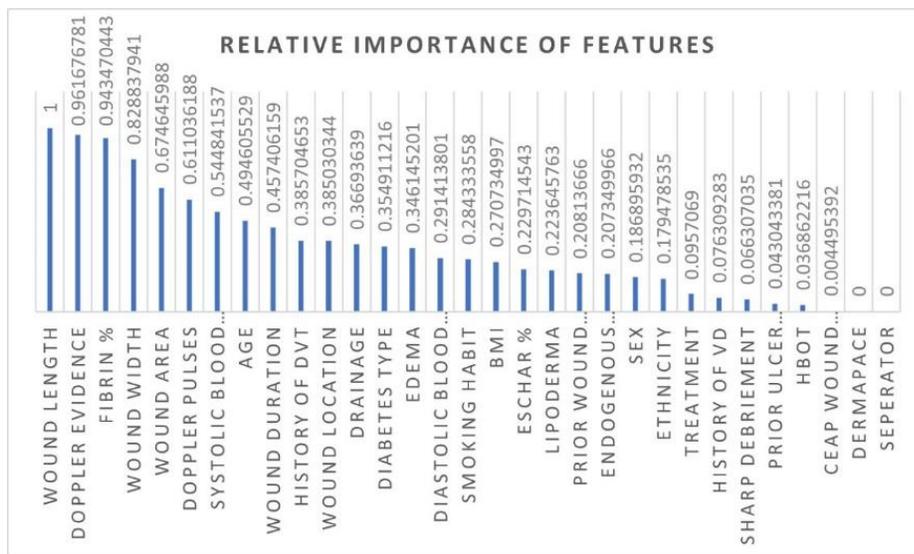

Figure 3- Relative importance of the prognosis variables

(2) **Jensen-Shannon (JS) divergence:** JS divergence is a simple method to measure the similarity between two probability distributions. If $w_1$ and $w_2$ be the weights of the probability distribution of $p_1$ and $p_2$, the generalization of Jensen-Shannon divergence is defined as (35):

$$JS(p_1, p_2) = H(w_1 p_1 + w_2 p_2) - w_1 H(p_1) - w_2 H(p_2) \qquad (2)$$

in which H is the Shannon entropy function. Considering the probability distribution of original data, $p_r$, and the same number of synthetic data, $p_s$, the JS divergence is calculated for all the prognosis factors to quantitatively compare the distribution of synthetic instances generated by EMR-TCWGAN to RME-CWGAN.

(3) **Prognosis classification accuracy:** The generated samples should be realistic and useful in real-life applications. Hence, a post-hoc Prog-CNN was trained on synthetic data and tested on real data (TSTR). To test if the EMR-TCWGAN is not copying the train data, the classification accuracy is reported on the test dataset. Moreover, since the number of the training data is limited, this accuracy can validate the hypothesis that the generated instances by EMR-TCWGAN can be useful in training a prognosis model and to make sure that the prognosis model will be accurate to predict healing status of new patients.

(4) **Discriminative accuracy:** The generator should produce samples that are indistinguishable from the real data. Therefore, a post-hoc classifier was trained to classify between real and fake samples. The classifier which is a CNN with two convolutional layers is trained on an equal number of real and synthetic instances and tested on the synthetic data. The classifier must classify a given instance as real or fake. For an excellent generator, the classifier should achieve less than 50% accuracy at this task (13).

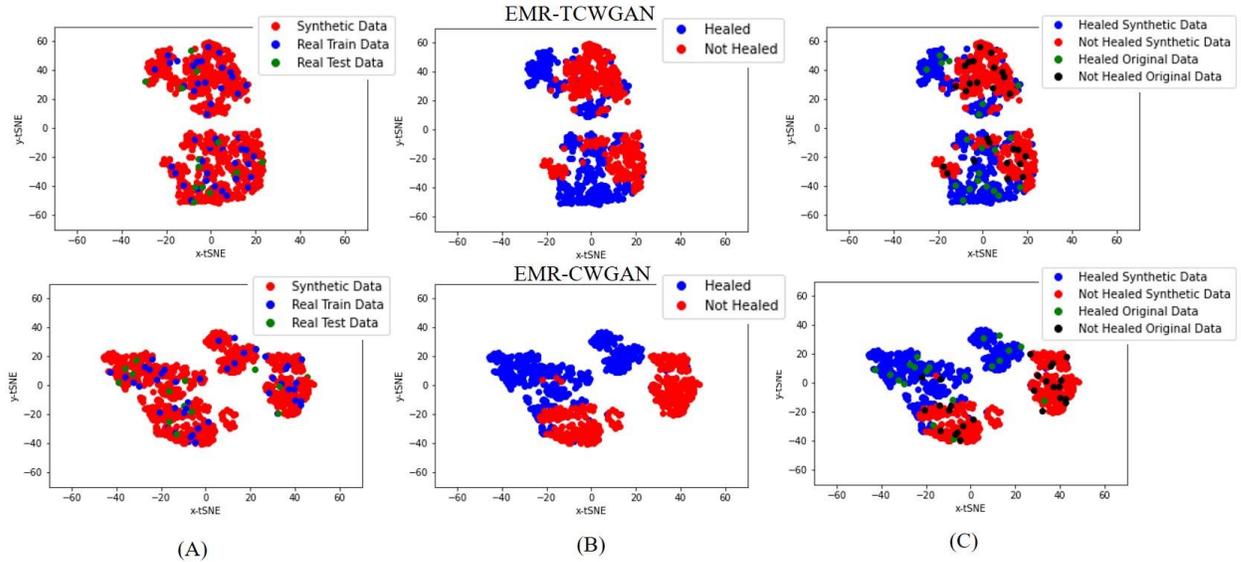

Figure 4- t-SNE visualization of timeseries EMR data generated by the proposed EMR-TCWGAN (first row) and the baseline model EMR-CWGAN (second row). (A) Synthetic and real data distribution, red denotes synthetic, blue denotes original train and green denotes original test data mapped into two dimensional. (B) healed vs. not healed distribution in the generated data. Red denotes generated data labeled as not healed, blue denotes generated data labeled as healed. (C) healed vs. not healed distribution in synthetic and real data. Red denotes generated data labeled as not healed, blue denotes generated data labeled as healed. Black denotes real data labeled as not healed. Green denotes real data labeled as healed.

Figure 4 illustrates the two-dimensional distribution of the synthetic (red dots), train (blue dots), and test (green dots) dataset. The dimension reduction was performed using t-SNE. In Figure 4B, the synthetic instances were categorized by their labels representing if a generated sample is healed by week 12 of the initial visit. The distribution of the labels in the synthetic samples was compared to the ground truth of the original dataset in Figure 4C. Comparing the distribution of the synthetic instances generated by EMR-TCWGAN (first row) to the ones generated by the baseline model (second row), a comparable performance, if not slightly better, is observed in samples generated by EMR-TCWGAN. However, it is seen that EMR-TCWGAN has captured the label distribution of the ground truth distribution better.

To quantitatively compare the distribution of the generated samples with the original dataset, The JS divergence was calculated for each prognosis factor and the average was reported in Table 2. Surprisingly, the JS divergence shows higher similarity between the distribution of the synthetic and the original dataset in all three follow-up visits when samples are generated by EMR-TCWGAN. Moreover, to evaluate the quality of the generated continuous features including wound length, wound width, and wound area, as well as their temporal variations, Figure 5 is illustrated to compare the distribution performance of the baseline model with EMR-TCWGAN. The temporal changes in EMR-TCWGAN show a better match with the temporal variations in the original data. Besides, JS divergence reported in Table 3 confirmed the fact that the proposed EMR-TCWGAN, with a few exceptions, outperformed the baseline model in capturing the time variations in wound length and wound area.

Table 2- The average JS divergence between the probability distribution function of all the prognosis factors in real and synthetic dataset generated by the proposed EMR-TCWGAN and the baseline model EMR-CWGAN. Bolds indicate the more similarity.

|  | T=1 | T=2 | T=3 |
|---|---|---|---|
| **EMR-TCWGAN** | **0.133** | **0.134** | **0.130** |
| **EMR-CWGAN** | 0.145 | 0.159 | 0.143 |

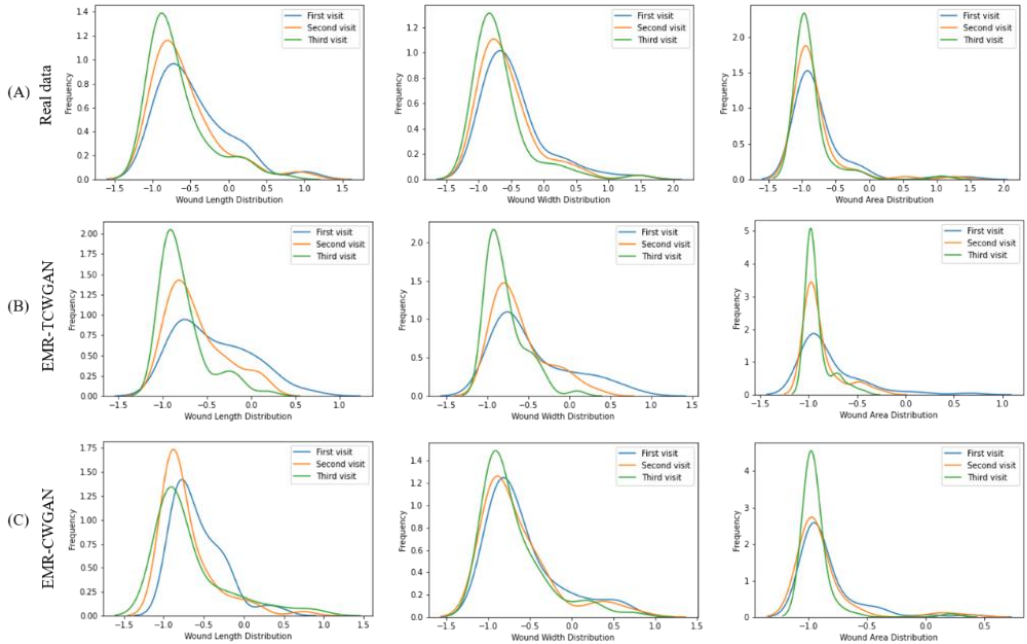

Figure 5- Probability density function of the continuous features (wound length, width, and area) for three successive visits in (A) Real data, (B) Synthetic data generated by the proposed EMR-TCWGAN, (C) Synthetic data generated by the baseline model EMR-WGAN.

Table 3- JS divergence between the probability distribution function of the continuous prognosis factors in real and synthetic dataset generated by (A) the proposed EMR-TCWGAN, (B) the baseline model EMR-CWGAN in three successive visits. Bolds indicate the better result.

**(A) EMR-TCWGAN**

| Visit / Features | Wound Length | Wound Width | Wound Area |
| --- | --- | --- | --- |
| T=1 | **0.057** | **0.147** | **0.134** |
| T=2 | **0.136** | 0.133 | **0.129** |
| T=3 | 0.173 | 0.192 | **0.108** |

**(B) EMR-CWGAN**

| Visit / Features | Wound Length | Wound Width | Wound Area |
| --- | --- | --- | --- |
| T=1 | 0.198 | 0.162 | 0.212 |
| T=2 | 0.158 | **0.094** | 0.215 |
| T=3 | **0.072** | **0.149** | 0.244 |

The prognosis classification accuracy and AUC of Prog-CNN are indicated in Table 4. The AUC measures how well a model can predict the outcome. A random guess results in AUC of 0.5, while a prefect model will achieve AUC of 1.0. Generally, models with AUC above 0.7 are considered as an acceptable predictive model fit (30). Prog-CNN was trained using synthetic instances generated by GANs and tested on the original data. T indicates the number of successive visits that are used in training the Prog-CNN. Generally, the prognosis model which is trained by the samples generated by EMR-TCWGAN shows higher AUC and classification accuracy compared to the one trained by fake instances coming from the EMR-CWGAN generator. Due to less temporal information in the network, the classification accuracy decreased from 89.99% to 83.33% and then to 80.00% when Prog-CNN network was trained using data generated by the proposed EMR-TCWGAN from the first three, first two, and first visits, respectively. Similarly, AUC reduced from 0.975 to 0.986 and then 0.849 when the temporal information has reduced from three visits to one visit.

Table 4- Predictive accuracy and the area under curve (AUC) of the prognosis model trained using data generated by EMR-TCWGAN and EMR-CWGAN. T indicates the number of the follow-up visits up. Bolds indicate the better results.

|                | EMR-TCWGAN | | EMR-CWGAN | |
|----------------|------------|-------|-----------|-------|
| Visits/metrics | Accuracy   | AUC   | Accuracy  | AUC   |
| T=1            | **80.00%** | **0.849** | 69.99% | 0.795 |
| T=2            | **83.33%** | **0.968** | 75.00% | 0.832 |
| T=3            | **89.99%** | **0.975** | 83.33% | 0.857 |

The discriminative accuracy is indicated in Table 5. Remarkably, it is observed that discriminative accuracy in samples generated by EMR-TCWGAN is relatively lower than the ones generated by EMR-CWGAN by 45.03%. As was mentioned before, the discriminative accuracy represents how well a classifier can distinguish between real and fake instances. The post-hoc classifier was tested only on the synthetic data; therefore, 23.97% discriminative accuracy means that 76.03% of the generated samples by EMR-TCWGAN were realistic enough to be classified as real by the post-hoc classifier mistakenly. However, this number was reduced to only 31% in samples generated by EMR-CWGAN. Broadly speaking, the suggested EMR-TCWGAN generates higher-quality and more realistic synthetic data in comparison to the baseline model.

Table 5- Discriminative accuracy of the post-hoc classifier to classify real vs. fake on samples generated by EMR-TCWGAN and EMR-CWGAN. Bold indicate the better result.

| metrics | EMR-TCWGAN | EMR-CWGAN |
|---------|------------|-----------|
| **Discriminative accuracy** (lower the better) | **23.97%** | 69.00% |

## Discussion

In this study, we used time-series EMR to predict if a VLU is healed within 12 weeks of the first initial intake exam. This study has several notable achievements for designing a prognosis model using a limited amount of data.

First, using GANs techniques in time series medical data provided a pathway to overcoming the challenges of accessing the electronic medical records of patients with VLU. This limited access is due to the privacy, security, and difficulties in the extraction of useful information.

Second, the new representation of EMR data based on the prognosis factors, not only has limited the input dimension, and therefore the need for an autoencoder to reduce the dimensionality, but also eliminated the need of having access to a big EMR dataset to train a robust medGAN.

Third, applying deep learning models (convolutional neural networks) in medical GANs enabled the ability to generate both categorical and continuous time-series data that can be useful in training a real-life model. Moreover, combining the conditional training with Wasserstein divergence allowed the medical GAN to generate more realistic EMR data when there is a limited number of train data, compared to the simple training strategy (16). The performance of EMR-TCWGAN in both prognosis classification accuracy and discriminative accuracy was relatively higher than the baseline model, suggesting that CNN will perform better than feedforward networks in generating EMR data.

Finally, training a prognosis model, as a real-life application for GAN, not only validates the strength of the suggested EMR-TCWGAN in producing synthetic time-series EMR but also, may help the clinicians with management decisions by alerting them early in the course of treatment if the VLU has a low probability of healing. Our prognosis model's performance is higher than previously published models. For instance, Kenneth Jung, et al, (31) reported AUCs between 0.834 and 0.847 using the data from the first and second (T=2) wound assessments by training logistic regression and random forests. However, in our model, as reported in Table 4, we have achieved the AUC of 0.968 when using the information of the first two visits by applying deep learning models. Using EMRs from a large EMR, Sang Kyu Cho et al. (24) achieved the AUC of 0.717 by training a classification decision tree using the patients' EMR after the first initial intake exams (T=1). However, our model achieved a significantly higher AUC, 0.849, when training the network using the data from the first initial intake exam.

Based on the definition of a chronic wound, one that has not healed in 4-12 weeks, it would be reasonable to consider the critical prognosis factors collected from the first three visits. Moreover, it was reported that although four weeks is a short period to predict a wound healing status clinically, in practice, a shorter time for prediction would be beneficial to help clinicians to decide on and modify treatment strategies (3). Our study suggests that it may not be feasible to accurately predict the wound healing rate using only the first week data (in this case we achieved AUC=0.849), we believe data from three weeks of follow-up will provide enough information to calculate a strong prediction of healing potential (with AUC=0.975). This early forecast and prediction of wound healing will enhance clinical outcomes and provide efficiencies in care.

## Conclusion

We improved the pipeline of the medical GANs by representing patients' EMR data based on their prognosis factors and applying deep learning models in medGAN to generate time-series continuous and categorical EMR data. We utilized samples generated by medGAN in training a real-world prognosis classifier to predict the wound healing status within 12 weeks of the first visit. Our experimental results illustrated that the proposed EMR-TCWGAN outperforms the previous EMR-GAN. Moreover, the prognosis accuracy has shown a promising result for clinical decision making.